\documentclass[11pt,a4paper]{article}

\usepackage[T1]{fontenc}
\usepackage[utf8]{inputenc}
\usepackage[margin=1in]{geometry}
\usepackage{amsmath,amssymb}
\usepackage{booktabs}
\usepackage{array}
\usepackage{multirow}
\usepackage{graphicx}
\usepackage{xcolor}
\usepackage{parskip}
\usepackage[colorlinks=true,linkcolor=blue,citecolor=blue,urlcolor=blue]{hyperref}
\usepackage{url}

\newcolumntype{L}[1]{>{\raggedright\arraybackslash}p{#1}}
\newcolumntype{C}[1]{>{\centering\arraybackslash}p{#1}}

\begin{document}

\title{\textbf{Please Make it Sound like Human: Encoder-Decoder vs.\
Decoder-Only Transformers for AI-to-Human Text Style Transfer}}

\author{
  Utsav Paneru \\
  Department of Computer Science, Kathmandu Engineering College, \\
  Kalimati, Kathmandu, Nepal \\
  \texttt{utsav.bct80095@kecktm.edu.np}
}

\date{}
\maketitle

\begin{abstract}
AI-generated text has become common in academic and professional writing,
prompting substantial research into detection methods. Far less attention has gone
to the reverse problem: can we systematically rewrite AI-generated prose so it
reads as genuinely human-authored? This paper addresses that question directly.

We build a parallel corpus of 25,140 paired AI-input and human-reference text
chunks from a multi-style writing repository, then perform a systematic linguistic
analysis identifying 11 measurable stylistic markers that separate the two
registers. The most striking difference is contraction usage: AI inputs average
0.00 contractions per chunk while human references average 0.17. Human text
also reads roughly four grade levels lower on the Flesch-Kincaid scale, uses
shorter words on average, and shows much greater sentence-length
variance---a texture that turns out to be surprisingly difficult to reproduce
automatically.

We fine-tune three models on this corpus: BART-base (139M parameters) and
BART-large (406M parameters) as encoder-decoder baselines, and
Mistral-7B-Instruct with QLoRA as a decoder-only alternative. The results push
back against the assumption that larger instruction-tuned models are inherently
better at style transfer. BART-large achieves the highest reference similarity
across all three overlap metrics---BERTScore F1 of 0.924, ROUGE-L of 0.566, and
chrF++ of 55.92---despite having 17 times fewer parameters than Mistral-7B. Its
outputs also converge onto the human distributional target with notable precision
on markers such as word length, lexical diversity, and readability. Mistral-7B
achieves a higher aggregate marker shift score, but per-marker inspection reveals
systematic overshoot rather than accurate targeting: the model overshoots the
human distribution on five markers and moves comma usage in the wrong direction
entirely. We argue that the gap between shift magnitude and shift accuracy is a
meaningful methodological blind spot in style transfer evaluation, and that
BART's denoising pretraining objective provides a structural advantage for
constrained rewriting that scale alone cannot compensate for.
\end{abstract}

\textbf{Keywords:} text style transfer, AI-to-human rewriting, seq2seq models,
BART, Mistral, linguistic markers, BERTScore, natural language generation

\section{Introduction}

Anyone who has used a large language model as a drafting aid has probably noticed
the same odd quality: the output is fluent and well-structured, but it doesn't read
like something a person wrote. The vocabulary tends toward the formal, sentences
follow each other with a characteristic smoothness, and the whole thing has a
slightly polished, impersonal quality that becomes easy to recognize after a while.
This isn't accidental. Language models are trained to produce coherent, fluent text,
and in doing so they develop consistent stylistic habits that differ systematically
from the variability and idiosyncrasy of actual human writing.

That divergence has driven a growing body of work on AI text detection
\cite{mitchell2023detectgpt,tang2023science}, which exploits exactly these
regularities. Much less studied is the other direction: building systems that learn
to close the stylistic gap, taking AI-generated text and rewriting it in a way that
genuinely reads as human-authored. We call this the \emph{AI-to-human style
transfer} problem.

The motivation is practical as well as scientific. Many people use AI as a drafting
assistant but want the final result to reflect their own voice. Researchers studying
AI's societal footprint need tools that can model the transformation between the two
registers. And building a humanizer forces a rigorous characterization of what human
writing actually looks like at a statistical level---one that turns out to be quite
informative in its own right.

Our approach is careful in execution if straightforward in design. We construct a
parallel corpus of AI-input and human-reference passages, quantify 11 stylistic
features that reliably distinguish the two styles, and fine-tune three model
configurations on this data. Critically, we evaluate not only whether outputs
preserve meaning---the standard measure---but also whether they actually shift those
11 markers toward the human distribution, and with what degree of accuracy.

The results are instructive. BART-large, at 406M parameters, consistently
outperforms Mistral-7B on every reference-based metric and produces outputs that
land much more precisely on the human distributional target. Mistral-7B's higher
mean marker shift, we show, reflects overshoot rather than accuracy: the model moves
past the human distribution on multiple markers instead of converging to it. This
distinction---between how far a model shifts the text and whether it shifts it to
the right place---is not well captured by current aggregate metrics, and we argue
it should be.

\subsection{Problem Statement}

We frame AI-to-human style transfer as conditional text generation. Given an
AI-generated passage $x$, we seek a model $f_\theta$ that produces a rewrite
$\hat{y}$ satisfying two constraints simultaneously: (i) \emph{semantic
preservation}---$\hat{y}$ must convey the same propositional content as $x$---and
(ii) \emph{stylistic transformation}---$\hat{y}$ must exhibit the distributional
properties of human-authored text across the 11 markers identified in
Section~\ref{sec:linguistic}. This dual criterion provides a concrete, quantitative
operationalization of what it means for text to sound human, moving beyond holistic
impressions to measurable linguistic dimensions.

\subsection{Contributions}

This paper makes the following contributions:

\begin{enumerate}
  \item We construct a parallel AI--human corpus of 25,140 chunk-level training
    pairs from a multi-style text repository, using document-disjoint
    train/validation/test splits to prevent leakage.
  \item We identify and quantify 11 linguistic markers that reliably differentiate
    AI-generated from human-authored text.
  \item We introduce the distinction between marker shift \emph{magnitude} and
    marker shift \emph{accuracy}, and show that aggregate shift scores can be
    genuinely misleading when models overshoot their target distribution.
  \item We present a controlled three-way comparison of BART-base, BART-large, and
    Mistral-7B on identical data, establishing that BART-large achieves superior
    performance with 17$\times$ fewer parameters.
  \item We argue that BART's denoising pretraining objective provides a structural
    advantage for style transfer that autoregressive pretraining at greater scale
    does not replicate.
\end{enumerate}

\section{Related Work}

\subsection{Detecting AI-Generated Text}

Work on detecting machine-generated text has grown substantially in recent years.
DetectGPT \cite{mitchell2023detectgpt} observes that AI-generated text tends to sit
at local maxima of the model's log-probability surface: small random perturbations
consistently reduce the probability of AI text but not of human text. Other
approaches train classifiers on stylistic features \cite{guo2023chatgpt} or embed
detectable watermarks at generation time \cite{tang2023science}. Our work is
complementary. Where detection methods exploit the statistical regularities that
distinguish AI from human writing, we are trying to eliminate those regularities.
Understanding what makes AI text detectable is therefore directly relevant to
building something that can close the stylistic gap through genuine transformation.

\subsection{Text Style Transfer}

Text style transfer modifies the stylistic register of a text while preserving its
semantic content \cite{jin2022deep}. Prior work has addressed formality transfer
\cite{rao2018dear}, sentiment reversal \cite{shen2017style}, and author imitation
\cite{wegmann2022author}. Most of these tasks involve shifting along a single
axis---from informal to formal, say, or from positive to negative sentiment. Our
task is more demanding: we need to shift simultaneously along 11 axes, moving from a
tightly clustered AI stylistic distribution toward the more variable and less
predictable distribution of human prose. To our knowledge, this specific multi-axis
formulation has not been studied in the style transfer literature.

\subsection{Encoder-Decoder Models for Generation}

BART \cite{lewis2020bart} is pretrained with a document corruption and
reconstruction objective, training the model to recover original text from a noisy
input. This denoising objective is structurally similar to style transfer: given a
stylistically ``corrupted'' (AI-generated) version of a passage, recover the human
original. We hypothesize that this alignment between pretraining objective and task
structure gives BART an inherent advantage for style transfer. T5 \cite{raffel2020t5},
the prior dominant seq2seq baseline for style transfer, was pretrained on a filtered
web corpus oriented toward task completion rather than stylistic fidelity, which may
help explain its weaker performance on such tasks in prior work.

\subsection{Decoder-Only Language Models}

The GPT family \cite{radford2019language} demonstrated that autoregressive models
trained on large corpora develop rich generative capabilities. LLaMA
\cite{touvron2023llama} extended this paradigm to open-weight models, and Mistral-7B
\cite{jiang2023mistral} improves on LLaMA-class models through grouped-query and
sliding-window attention, achieving strong benchmark performance at the 7B scale. We
use Mistral-7B-Instruct-v0.2 because it combines strong language modeling with
instruction-following capability, making it a reasonable candidate for constrained
rewriting.

\subsection{Parameter-Efficient Fine-Tuning}

QLoRA \cite{dettmers2023qlora} makes large-model fine-tuning tractable on consumer
hardware by combining 4-bit NF4 quantization of base model weights with low-rank
adapter modules as the only trained parameters. This enables 7B-scale fine-tuning on
a single GPU at some cost in representational fidelity relative to full-precision
training.

\section{Dataset Construction}

\subsection{Source Corpus}

We draw human-authored text from a curated Hugging Face repository spanning
multiple writing styles and domains---formal academic prose, technical
documentation, and creative writing. This stylistic breadth is deliberate. Corpora
restricted to a single domain tend to produce humanizers that learn domain-specific
surface patterns rather than general properties of human prose. Broader coverage is
the more principled choice.

From this corpus we sample passages and generate semantically equivalent AI-style
rewrites using two LLaMA-family models: llama-3.3-70b-versatile and
llama-3.1-8b-instant, with the generating model recorded per example in the metadata
(fields: doc\_id, chunk\_idx, ai, human, style, model, prompt\_id). Using two
generators rather than one reduces the risk that a trained humanizer learns to undo
one model's idiosyncratic habits rather than AI-style writing more broadly. The
generation prompt instructs each model to preserve meaning while adopting formal,
structured AI-like phrasing and avoiding contractions, slang, and casual
expressions.

\subsection{Sentence-Aware Chunking}

Source passages frequently exceed the context window of smaller models. We address
this with sentence-aware chunking: passages are segmented using NLTK sentence
tokenization, then sentences are greedily grouped into chunks of at most 200 tokens
as measured by the BART-base tokenizer. Chunk pairs where either side falls below 10
words are discarded. Human and AI chunks are aligned positionally within each
document, so chunk $i$ of the AI version corresponds to chunk $i$ of the human
version.

This process yields 25,140 training chunk pairs, with an additional 1,390 validation
and 1,390 test examples. Splits are document-disjoint: every chunk from a given
source document appears in exactly one split, eliminating cross-contamination between
sets. Chunking also multiplies usable training data and provides a cleaner learning
signal---rewriting a 140-word chunk is a more tractable problem than rewriting a
400-word essay.

\subsection{Linguistic Analysis}
\label{sec:linguistic}

Before training, we characterized the stylistic gap between human and AI text by
measuring 11 features across the test subset ($n = 1{,}390$). Results appear in
Table~\ref{tab:markers}. The most discriminative feature is contraction usage: AI
inputs average 0.00 contractions per chunk, human references average 0.17. Human
chunks also score far lower on the Flesch-Kincaid Grade Level (11.5 vs.\ 17.8), read
more easily on the Flesch Reading Ease scale (46.1 vs.\ 14.1), and show roughly
twice the sentence-length variance of AI text (37.1 vs.\ 18.4). AI chunks use longer
words (5.84 vs.\ 5.09 characters) and higher lexical diversity (0.853 vs.\ 0.783),
consistent with the formal, varied vocabulary typical of AI-generated prose. These
patterns hold across the full training corpus.

\begin{table}[ht]
\centering
\caption{Chunk-level linguistic marker comparison between human-authored and
AI-generated text (test subset, $n = 1{,}390$).}
\label{tab:markers}
\small
\begin{tabular}{lrrrc}
\toprule
\textbf{Metric} & \textbf{Human} & \textbf{AI} & \textbf{Change} & \textbf{Signal} \\
\midrule
Avg word count          & 50.77 & 41.84 & $-17.5\%$ & Moderate \\
Avg sentence count      & 3.78  & 2.41  & $-36.2\%$ & Moderate \\
Avg word length (chars) & 5.09  & 5.84  & $+14.7\%$ & Strong \\
Lexical diversity       & 0.783 & 0.853 & $+8.9\%$  & Moderate \\
Contractions per chunk  & 0.17  & 0.00  & $-100\%$  & Very Strong \\
Question marks          & 0.084 & 0.035 & $-58.3\%$ & Strong \\
Exclamations            & 0.065 & 0.013 & $-80.0\%$ & Strong \\
Commas per chunk        & 2.66  & 3.38  & $+27.1\%$ & Strong \\
Sentence length variance& 37.1  & 18.4  & $-50.4\%$ & Strong \\
Flesch Reading Ease     & 46.1  & 14.1  & $-69.4\%$ & Very Strong \\
Flesch-Kincaid Grade    & 11.5  & 17.8  & $+54.8\%$ & Very Strong \\
\bottomrule
\end{tabular}
\end{table}

\section{Methodology}

\subsection{Problem Formulation}

We treat AI-to-human style transfer as conditional text generation. Given an
AI-generated chunk $x$, we seek a model $f_\theta$ producing a rewrite
$\hat{y} = \arg\max_y P_\theta(y \mid x)$ that satisfies two constraints: (1)
$\hat{y}$ preserves the semantic content of $x$, evaluated via BERTScore
\cite{zhang2019bertscore}; and (2) $\hat{y}$ exhibits the distributional properties
of human-authored text across the 11 linguistic markers, evaluated via the marker
shift framework described in Section~\ref{sec:eval}.

\subsection{Model Configurations}

We evaluate three model configurations spanning two architectural families.

\subsubsection{BART-base}

BART-base (\texttt{facebook/bart-base}) is a 139M-parameter encoder-decoder model
pretrained with a denoising objective on BookCorpus and Wikipedia
\cite{lewis2020bart}. We prepend the task prefix \texttt{``humanize: ''} to each AI
chunk and train the model to output the corresponding human chunk using standard
cross-entropy loss over the full target sequence.

Training is full fine-tuning in bf16 precision, learning rate $5\times10^{-5}$ with
a linear schedule and 5\% warmup, effective batch size 16 (4 per device $\times$ 4
gradient accumulation steps).

\textit{Important caveat:} the saved training configuration reflects a limited
smoke-test run (\texttt{max\_steps=10}, \texttt{max\_train\_samples=128}, 1 epoch)
used only to verify the pipeline end-to-end. BART-base results should be interpreted
as a constrained lower-bound baseline rather than a fully trained model.

\subsubsection{BART-large}

BART-large (\texttt{facebook/bart-large}) scales the same encoder-decoder
architecture to 406M parameters with wider hidden dimensions and more attention
heads. Task format and training objective are identical to BART-base. Full
fine-tuning in bf16 precision; learning rate $5\times10^{-5}$ with a cosine
schedule and 10\% warmup; effective batch size 16 (2 per device $\times$ 8 gradient
accumulation steps); 5 epochs with per-epoch evaluation and best-checkpoint
selection based on validation loss.

\subsubsection{Mistral-7B with QLoRA}

Mistral-7B-Instruct-v0.2 is a 7B-parameter decoder-only model. We attach LoRA
adapters (rank $r = 16$, scaling $\alpha = 32$, dropout 0.05) to the query, key,
value, and output projection layers of each attention block, yielding approximately
80M trainable parameters against the frozen 4-bit NF4 quantized
base---roughly 1.1\% of total parameters.

Training uses QLoRA-style 4-bit NF4 quantization with double quantization and
float16 compute dtype, in fp16 precision with the Paged AdamW 32-bit optimizer.
Learning rate $2\times10^{-4}$ with a cosine schedule and 5\% warmup; effective
batch size 8 (2 per device $\times$ 4 gradient accumulation steps); maximum 500
steps with checkpoints every 100 steps. We apply completion-only loss masking using
the \texttt{``\#\#\# Response:''} delimiter so the model is trained only on the
target rewrite, not the instruction prefix. Inputs are formatted as:

\begin{verbatim}
### Instruction: [rewrite instruction]
### Input: {ai_text}
### Response: {human_text}
\end{verbatim}

\begin{table}[ht]
\centering
\caption{Training configuration summary.}
\label{tab:training}
\small
\begin{tabular}{lccc}
\toprule
\textbf{Setting} & \textbf{BART-base} & \textbf{BART-large} & \textbf{Mistral-7B} \\
\midrule
Parameters      & 139M (full FT)  & 406M (full FT)  & 7B (LoRA only) \\
Adapter         & None            & None            & QLoRA NF4, $r$=16, $\alpha$=32 \\
Epochs / Steps  & 1 epoch (smoke) & 5 epochs        & 500 steps \\
Learning rate   & $5\times10^{-5}$ (linear) & $5\times10^{-5}$ (cosine) & $2\times10^{-4}$ (cosine) \\
Batch size (eff.)& 16             & 16              & 8 \\
Precision       & bf16            & bf16            & fp16 + 4-bit base \\
Optimizer       & AdamW           & AdamW           & Paged AdamW 32-bit \\
\bottomrule
\end{tabular}
\end{table}

\section{Evaluation Framework}
\label{sec:eval}

We evaluate along five complementary dimensions.

\textbf{BERTScore} \cite{zhang2019bertscore} measures semantic similarity between
model output and human reference using contextual BERT embeddings. We report
precision, recall, and F1. High BERTScore combined with successful marker shift is
the joint criterion for successful humanization.

\textbf{ROUGE-L and chrF++} are lexical overlap metrics capturing surface-level
fidelity. ROUGE-L operates via longest common subsequence; chrF++ works at the
character $n$-gram level and is more robust to morphological variation. Both provide
a view complementary to embedding-based similarity.

\textbf{GPT-2 Perplexity} is an indirect proxy for naturalness: we compute the
perplexity of model outputs under a GPT-2 language model. Human reference text
scores 23.69. Outputs substantially above this value appear less fluent than human
writing; outputs substantially below it may be over-smooth and predictable rather
than genuinely natural. We interpret this metric cautiously.

\textbf{Vocabulary Jaccard} measures lexical overlap between model outputs and human
references at the type level, providing a complementary view of vocabulary
alignment.

\textbf{Linguistic Marker Shift} is our primary novel evaluation framework. For each
of the 11 markers, we compute a directional shift score capturing how far the output
has moved from the AI input value toward the human reference mean. Formally:
\[
  \text{shift} = \frac{\text{output} - \text{AI}}{\text{human} - \text{AI}},
  \quad \text{clipped to } [-1, 2]
\]
A score of 1.0 means the output exactly matches the human mean; values above 1.0
signal overshoot; values below 0 indicate movement in the wrong direction. We report
per-marker shifts and the mean across all 11 markers.

\section{Results}

\subsection{Reference Similarity}

\begin{figure}[ht]
  \centering
  \includegraphics[width=0.85\textwidth]{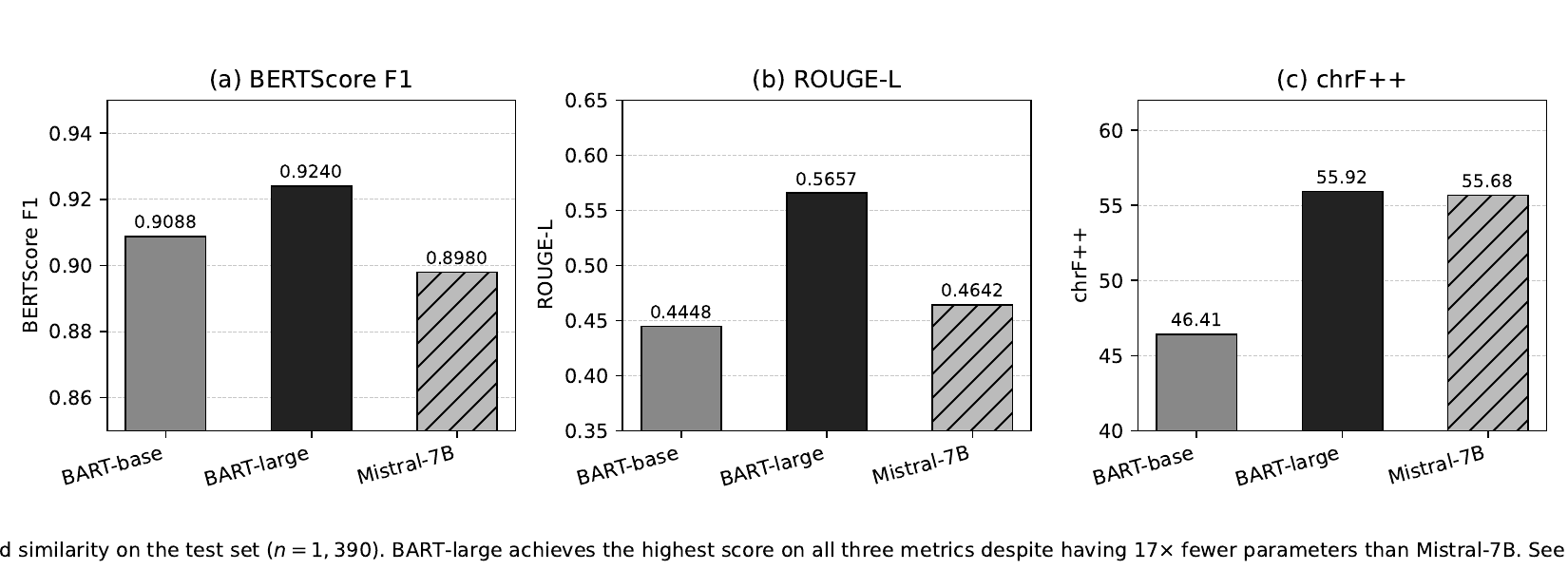}
  \caption{Reference-based metrics (BERTScore F1, ROUGE-L, chrF++) for BART-base,
  BART-large, and Mistral-7B on the 1,390-example test set.}
  \label{fig:metrics}
\end{figure}

Table~\ref{tab:results} presents quantitative results across all models, and
Figure~\ref{fig:metrics} visualizes the three reference-based metrics side by side.
The pattern is consistent: BART-large outperforms both BART-base and Mistral-7B on
BERTScore F1 (0.924 vs.\ 0.909 and 0.898), ROUGE-L (0.566 vs.\ 0.445 and 0.464),
and chrF++ (55.92 vs.\ 46.41 and 55.68). The consistency across all three
metrics---which differ substantially in what they measure, from contextual semantic
similarity to character-level fidelity---suggests that BART-large's advantage is not
an artifact of any single metric's properties.

The advantage over Mistral-7B is most pronounced on ROUGE-L ($+10.1$ percentage
points) and modest but consistent on BERTScore F1 ($+2.6$ points). That BART-large
achieves this with 406M parameters against Mistral-7B's 7B---a 17-fold
difference---is the central empirical finding of this paper.

\begin{table}[ht]
\centering
\caption{Quantitative evaluation results on the 1,390-example test set. Best values
per metric in bold. PPL = GPT-2 perplexity; Shift = mean linguistic marker shift.}
\label{tab:results}
\small
\begin{tabular}{lcccccc}
\toprule
\textbf{Model} & \textbf{Params} & \textbf{BERTScore F1} & \textbf{ROUGE-L} & \textbf{chrF++} & \textbf{PPL} & \textbf{Mean Shift} \\
\midrule
BART-base  & 139M & 0.9088          & 0.4448          & 46.41          & 26.69 & 0.6513 \\
BART-large & 406M & \textbf{0.9240} & \textbf{0.5657} & \textbf{55.92} & 27.15 & 0.8289 \\
Mistral-7B & 7B   & 0.8980          & 0.4642          & 55.68          & \phantom{0}9.03 & 1.2788 \\
Human ref  & ---  & ---             & ---             & ---            & 23.69 & --- \\
\bottomrule
\end{tabular}
\end{table}

\subsection{Linguistic Marker Shift: Magnitude vs.\ Accuracy}

Table~\ref{tab:markers_per} reports per-marker averages and Table~\ref{tab:shifts}
reports directional shift scores. Mistral-7B achieves the highest mean shift (1.279)
compared to BART-large (0.829) and BART-base (0.651). In isolation, this would
suggest Mistral-7B is the superior humanizer. The per-marker breakdown explains why
that conclusion is wrong.

Five of Mistral-7B's 11 marker shifts are capped at the maximum value of 2.0,
meaning the model's outputs have overshot the human distribution by as far past the
target as the AI input was before it. Word count in Mistral-7B outputs averages 77.9
per chunk, against a human target of 50.8---a 53\% overshoot. Sentence count
reaches 5.84 versus a human average of 3.78. Contraction usage reaches 0.383 per
chunk, more than double the human average of 0.17. Lexical diversity falls to 0.510,
well below the human average of 0.783. In each case, the model has traded one form
of non-human text for another---not AI-formal, but AI-verbose and AI-colloquial.

Most telling is the comma result: Mistral-7B's comma shift score is $-1.0$, meaning
it moves comma usage in the wrong direction entirely. AI text already uses more
commas than human writing (3.38 vs.\ 2.66 per chunk); Mistral-7B increases this
further to 4.43. BART-large, by contrast, reduces comma usage to 2.58---within 0.08
of the human average.

BART-large's outputs land precisely on the human distributional target for the
markers most predictive of perceived naturalness. Average word length in BART-large
outputs is 5.094 characters; the human reference mean is 5.094. Lexical diversity is
0.783 in both. Flesch Reading Ease reaches 44.8 against a human target of 46.1.
This precision is not captured by the mean shift score, and it represents a
qualitatively different kind of success from Mistral-7B's large-but-inaccurate
movements.

\begin{table}[ht]
\centering
\caption{Per-marker averages across AI input, model outputs, and human reference.
Key: $\bigtriangleup$ = Mistral-7B overshoots the human target (shift $> 1$).
(WD) = wrong direction: Mistral-7B shift $< 0$, meaning the output moves away from
the human mean rather than toward it.}
\label{tab:markers_per}
\small
\begin{tabular}{lccccc L{1.6cm}}
\toprule
\textbf{Marker} & \textbf{AI Input} & \textbf{BART-base} & \textbf{BART-large} & \textbf{Mistral-7B} & \textbf{Human Ref} & \textbf{Note} \\
\midrule
Word count            & 41.8 & 48.9 & 46.3 & 77.9 & 50.8 & $\bigtriangleup$ \\
Sentence count        & 2.41 & 3.53 & 3.41 & 5.84 & 3.78 & $\bigtriangleup$ \\
Avg word length       & 5.84 & 5.52 & 5.09 & 5.16 & 5.09 & \\
Lexical diversity     & 0.853& 0.766& 0.783& 0.510& 0.783& $\bigtriangleup$ \\
Contractions          & 0.00 & 0.095& 0.097& 0.383& 0.166& $\bigtriangleup$ \\
Question marks        & 0.035& 0.047& 0.048& 0.112& 0.084& \\
Exclamations          & 0.013& 0.027& 0.037& 0.049& 0.065& \\
Commas per chunk      & 3.38 & 2.91 & 2.58 & 4.43 & 2.66 & (WD) \\
Sent.\ length var.    & 18.4 & 37.5 & 46.9 & 56.5 & 37.1 & $\bigtriangleup$ \\
Flesch Reading Ease   & 14.1 & 30.2 & 44.8 & 44.1 & 46.1 & \\
F-K Grade Level       & 17.8 & 14.0 & 11.7 & 11.7 & 11.5 & \\
\bottomrule
\end{tabular}
\end{table}

\begin{table}[ht]
\centering
\caption{Directional marker shift scores. Values $> 1$ indicate overshoot; values
$< 0$ indicate movement in the wrong direction.}
\label{tab:shifts}
\small
\begin{tabular}{lccc L{3.4cm}}
\toprule
\textbf{Marker} & \textbf{BART-base} & \textbf{BART-large} & \textbf{Mistral-7B} & \textbf{Note} \\
\midrule
Word count           & 0.785 & 0.496 & 2.000 & Mistral overshoots \\
Sentence count       & 0.823 & 0.730 & 2.000 & Mistral overshoots \\
Avg word length      & 0.438 & 1.000 & 0.908 & BART-large exact \\
Lexical diversity    & 1.245 & 1.004 & 2.000 & Mistral overshoots \\
Contractions         & 0.571 & 0.584 & 2.000 & Mistral overshoots \\
Question marks       & 0.246 & 0.275 & 1.565 & \\
Exclamations         & 0.278 & 0.458 & 0.694 & \\
Commas per chunk     & 0.653 & 1.116 & $-1.000$ & Mistral wrong direction \\
Sent.\ length var.   & 1.024 & 1.528 & 2.000 & Mistral overshoots \\
Flesch Reading Ease  & 0.503 & 0.962 & 0.937 & BART-large near exact \\
F-K Grade Level      & 0.598 & 0.965 & 0.962 & \\
\midrule
\textbf{Mean}        & 0.651 & 0.829 & 1.279 & \\
\bottomrule
\end{tabular}
\end{table}

\begin{figure}[ht]
  \centering
  \includegraphics[width=0.85\textwidth]{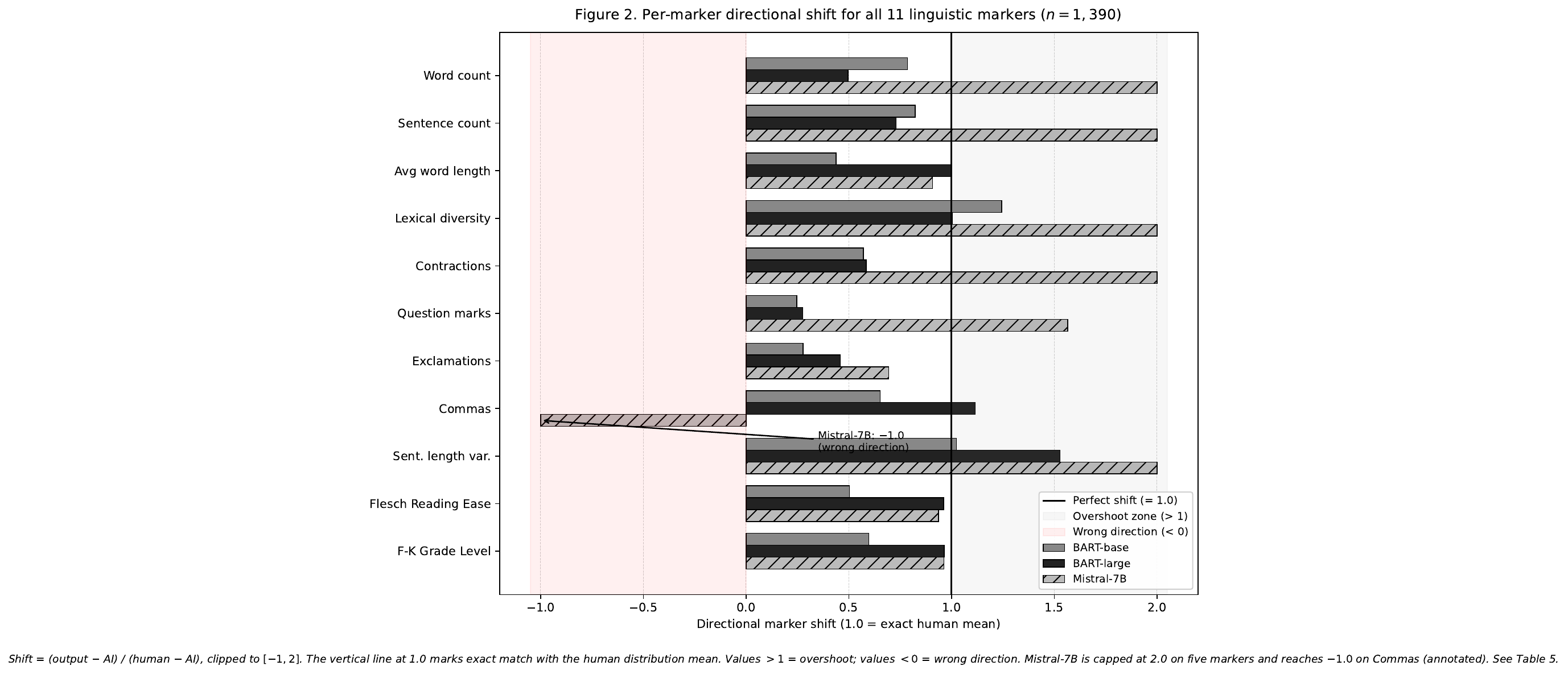}
  \caption{Per-marker directional shift scores for all three models. Values above
  1.0 indicate overshoot beyond the human target; values below 0 indicate movement
  in the wrong direction.}
  \label{fig:shifts}
\end{figure}

Note that BART-large itself overshoots on sentence length variance (shift = 1.528),
which the mean shift score does not highlight. This is a useful reminder that no
model achieves perfect accuracy across all markers. Figure~\ref{fig:shifts} plots
all 11 per-marker shift scores for each model together, making the contrast between
BART-large's distributional precision and Mistral-7B's systematic overshoot
immediately visible.

\subsection{Fluency: GPT-2 Perplexity}

\begin{figure}[ht]
  \centering
  \includegraphics[width=0.55\textwidth]{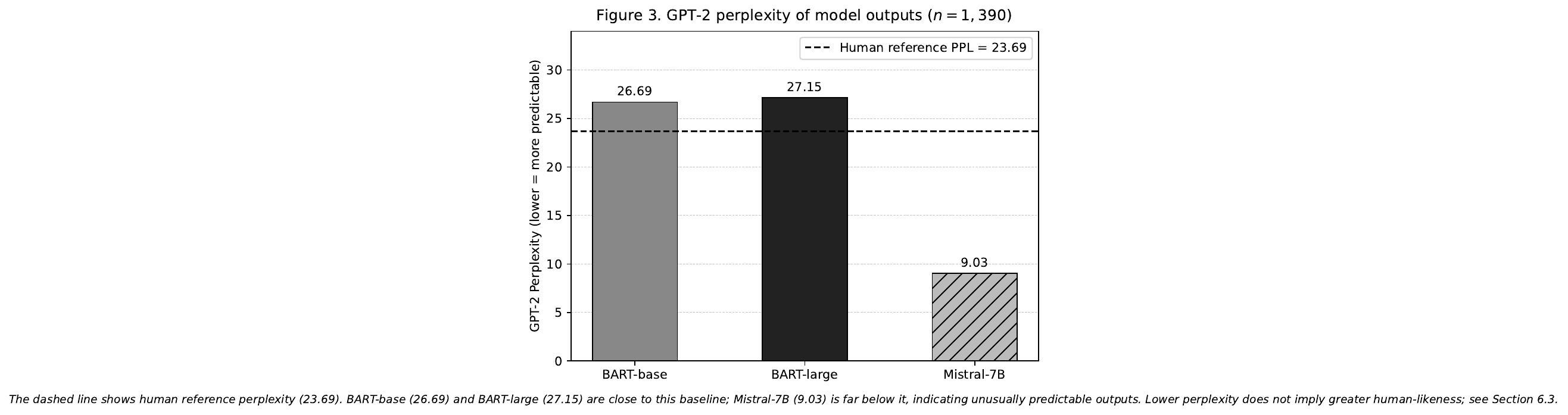}
  \caption{GPT-2 perplexity for model outputs and the human reference baseline
  (23.69). Lower is not always better: Mistral-7B's score of 9.03 reflects
  over-smooth generation rather than human-like naturalness.}
  \label{fig:ppl}
\end{figure}

Human reference text scores 23.69 under GPT-2 perplexity. BART-base (26.69) and
BART-large (27.15) both produce outputs close to this value, consistent with fluent
and naturalistic generation. Mistral-7B scores 9.03---dramatically lower than even
the human reference. Figure~\ref{fig:ppl} plots these values alongside the human
reference baseline.

This is counterintuitive but interpretable. Mistral-7B's outputs are so predictable
that a language model trained on human writing assigns them extremely low perplexity.
Human prose is inherently variable and occasionally surprising; text that is far more
predictable than genuine human writing has overcorrected toward fluency rather than
authenticity. It is worth noting that very low GPT-2 perplexity may also partly
reflect differences in register or sentence structure that make Mistral-7B outputs
more grammatically stereotypical rather than genuinely unnatural. The perplexity
result should be interpreted alongside the marker analysis rather than in isolation.
Taken together, both analyses point in the same direction: Mistral-7B produces text
that is confidently non-AI in register but not genuinely human in statistical
texture.

\section{Discussion}

\subsection{Why BART-large Outperforms Mistral-7B}

We attribute BART-large's superior reference similarity and distributional precision
to two complementary factors.

The first is architectural. BART's encoder-decoder design explicitly encodes the
full source text before any decoding begins, allowing the model to attend to the
complete AI input at every generation step. For a task requiring close tracking of
source content while modifying surface form, this is a meaningful structural
advantage over decoder-only models that compress input and output into a single
left-to-right pass.

The second, which we consider more fundamental, is that BART's denoising pretraining
objective is structurally isomorphic to style transfer. Given a stylistically
``corrupted'' (AI-generated) version of a passage, BART must recover the human
original. Fine-tuning for this task is therefore a direct instantiation of what the
model was pretrained to do: it requires learning a new definition of
``corruption,'' not a wholly new capability. Mistral-7B was pretrained with a
standard autoregressive objective and then instruction-tuned, neither of which
directly develops the constrained reconstruction capacity that style transfer
demands.

This interpretation suggests a practical principle: for style transfer tasks where a
reference corpus exists, models whose pretraining involves reconstruction under
constraints will tend to outperform models of greater raw capacity whose pretraining
does not.

\subsection{The Overshoot Problem and Evaluation Reform}

The most important methodological contribution of this paper is the distinction
between shift magnitude and shift accuracy. Mistral-7B's mean marker shift of 1.279
exceeds BART-large's 0.829, but this reflects how far the model moves, not whether
it moves in the right direction or stops at the right place. A model that overshoots
every marker by exactly the same amount it moved from the AI input would achieve a
mean shift of 2.0---the theoretical maximum---while producing text no more human than
the AI original, only differently non-human.

We propose that future work on AI-to-human style transfer report both directional
shift scores and absolute distributional distance---the mean absolute deviation
between output and human marker averages---to give a complete picture of stylistic
transformation quality. Under an absolute distance metric, BART-large's precise
landing on the human distribution would be clearly distinguished from Mistral-7B's
consistent overshoot.

\subsection{Implications for Scale and Architecture}

Our findings complicate the assumption that larger decoder-only models are
universally better at generation tasks. BART-large outperforms Mistral-7B on every
reference-based metric with 17 times fewer parameters, no quantization overhead, and
no adapter layers. For practitioners building humanization systems under resource
constraints---on-device inference, edge deployment, or cost-sensitive API
settings---this suggests that architecture and pretraining objective deserve more
weight as design considerations than raw scale.

This should not be read as a blanket argument against decoder-only models. For
open-ended generation tasks with no reference target, autoregressive models may well
be preferable. But for constrained rewriting where reference fidelity and
distributional precision matter, encoder-decoder models with
reconstruction-oriented pretraining represent a strong and resource-efficient
alternative.

\subsection{Limitations}

Several limitations bound how far these conclusions generalize.

Our AI-generation corpus was produced by LLaMA-family models. Different generators
exhibit somewhat different stylistic signatures, and a humanizer trained on this
corpus may overfit to LLaMA-specific patterns rather than AI-style writing in
general. Extending training to a multi-model generation corpus would reduce this
risk.

Our 11-marker linguistic framework does not capture all dimensions of writing
naturalness. Discourse coherence, pragmatic appropriateness, idiomatic expression,
and narrative voice are absent. It is plausible that Mistral-7B scores better on
dimensions we do not measure---the GPT-2 perplexity result suggests its outputs have
a fluency our marker analysis does not fully credit.

We do not include a human preference study, leaving open whether readers actually
perceive BART-large's outputs as more natural. Human judgments may diverge from
distributional analysis in ways that would complicate our conclusions.

We also do not test outputs against deployed AI detection systems. Demonstrating
that outputs evade detection while preserving meaning would provide strong practical
evidence that the stylistic shift is genuine.

Finally, BART-base was trained as a smoke-test rather than a fully trained model.
Its results establish a lower bound and should not be treated as a fair comparison to
BART-large or Mistral-7B on equal training terms.

\section{Conclusion}

We have compared encoder-decoder and decoder-only architectures for AI-to-human
text style transfer on a parallel corpus of 25,140 training chunk pairs and a
1,390-example test set, evaluating BART-base, BART-large, and Mistral-7B.

The central finding is that BART-large achieves superior performance on every
reference-based metric---BERTScore F1 of 0.924, ROUGE-L of 0.566, chrF++ of
55.92---while producing outputs that land with notable precision on the human
distributional target. It does this with 406 million parameters, 17 times fewer than
Mistral-7B.

Mistral-7B achieves a higher mean marker shift, but this reflects overshoot rather
than accuracy. The model moves past the human distribution on five markers and moves
comma usage in the wrong direction entirely. The gap between how much a model changes
the text and how accurately it targets the human distribution is not well captured by
current evaluation practice; we propose that absolute distributional distance be
reported alongside directional shift scores going forward.

The explanation we favor is structural: BART's denoising pretraining objective
directly develops the reconstruction-under-constraints capability that style transfer
requires, while Mistral-7B's autoregressive pretraining and instruction tuning do
not. For constrained rewriting tasks, the alignment between pretraining objective and
task structure appears to matter more than raw parameter count.

We hope this paper contributes both a practical finding---that efficient
encoder-decoder models remain genuinely competitive for style transfer despite the
field's drift toward large autoregressive models---and a methodological one:
evaluating style transfer requires measuring not just how far a model moves the text,
but whether it moves it to the right place.

\bibliographystyle{unsrt}
\bibliography{refs}

\end{document}